\documentclass{article}

\usepackage{microtype}
\usepackage{graphicx}
\usepackage{subfigure}
\usepackage{booktabs} 
\usepackage{amsmath}

\usepackage{hyperref}
\usepackage{gensymb}

\usepackage[accepted]{preprint-accepted-waise}
	
\icmltitlerunning{Improving Robustness of DNNs for Aerial Navigation by Incorporating Input Uncertainty}

\begin{document}

\twocolumn[
\icmltitle{Improving Robustness of Deep Neural Networks
for Aerial Navigation by Incorporating Input
Uncertainty}



\icmlsetsymbol{equal}{*}

\begin{icmlauthorlist}
\icmlauthor{Fabio Arnez}{CEA-LIST}
\icmlauthor{Huascar Espinoza}{CEA-LIST}
\icmlauthor{Ansgar Radermacher}{CEA-LIST}
\icmlauthor{Fran\c{c}ois Terrier}{CEA-LIST}
\end{icmlauthorlist}

\icmlaffiliation{CEA-LIST}{Universit\'{e} Paris-Saclay, CEA, List, F-91120, Palaiseau, France}

\icmlcorrespondingauthor{Fabio Arnez}{fabio.arnez@cea.fr}

\icmlkeywords{Machine Learning, ICML}

\vskip 0.3in
]



\printAffiliationsAndNotice{}  

\begin{abstract}
Uncertainty quantification methods are required in autonomous systems that include deep learning (DL) components to assess the confidence of their estimations. However, to successfully deploy DL components in safety-critical autonomous systems, they should also handle uncertainty at the input rather than only at the output of the DL components. Considering a probability distribution in the input enables the propagation of uncertainty through different components to provide a representative measure of the overall system uncertainty. In this position paper, we propose a method to account for uncertainty at the input of Bayesian Deep Learning control policies for Aerial Navigation. Our early experiments show that the proposed method improves the robustness of the navigation policy in Out-of-Distribution (OoD) scenarios.
\end{abstract}

\section{Introduction}
\label{Intro}
Autonomous navigation in complex environments still represents a big challenge for autonomous systems. Particular instances of this problem are autonomous driving for self-driving cars and autonomous aerial navigation in the context of Unmanned Aerial Vehicles (UAVs). In both cases, the navigation task is addressed by first acquiring rich and complex raw sensory information (e.g., from cameras, radars, LiDARs, etc.), which is then processed to drive the robot towards its goal. Usually, this process is done in a modular fashion, where specific software components are linked together in the so-called \textit{perception-planning-control} software pipeline \cite{siegwart2011introduction,mcallister2017concrete}.

Modular pipeline components can be implemented using Deep Neural Networks (DNNs) or other non-learning methods \cite{grigorescu2020survey}. DNNs have become a popular choice thanks to their effectiveness in processing complex sensory inputs, and their powerful representation learning, that surpass the performance of traditional methods. Navigation policies implemented in a modular fashion admit a higher degree of interpretability and ease of analysis since components are developed in isolation. However, these architectures suffer from the accumulation of errors (coming from the contribution of each component's erroneous outputs), which later harms the overall system performance \cite{mcallister2017concrete,mueller2018driving}. An alternative approach to modular pipelines is to map sensory inputs directly to control outputs using neural networks in an End-to-End (E2E) fashion \cite{grigorescu2020survey}. This is an appealing paradigm where perception-planning-control blocks are trained jointly, often via imitation learning \cite{codevilla2018end,codevilla2019exploring}. Unfortunately, E2E training requires vast amounts of data \cite{mcallister2017concrete,mueller2018driving} limiting its use to constrained scenarios.

Recently, some works from \cite{mueller2018driving,bonatti2019learning,wu2020towards} have explored the combination of modular and E2E learning approaches to get the benefits of both families. The goal of this hybrid approach is to learn an environment representation and a driving policy through dedicated DNNs. In this manner, an autonomous system can incorporate DL-based modules for perception and control. Learning intermediate (compact) representations enables the extraction of relevant environmental features that can be used later in downstream tasks (e.g. control policy training).

\begin{figure*}[ht!]
	\vskip 0.2in
	\begin{center}
		\centerline{\includegraphics[width=\textwidth]{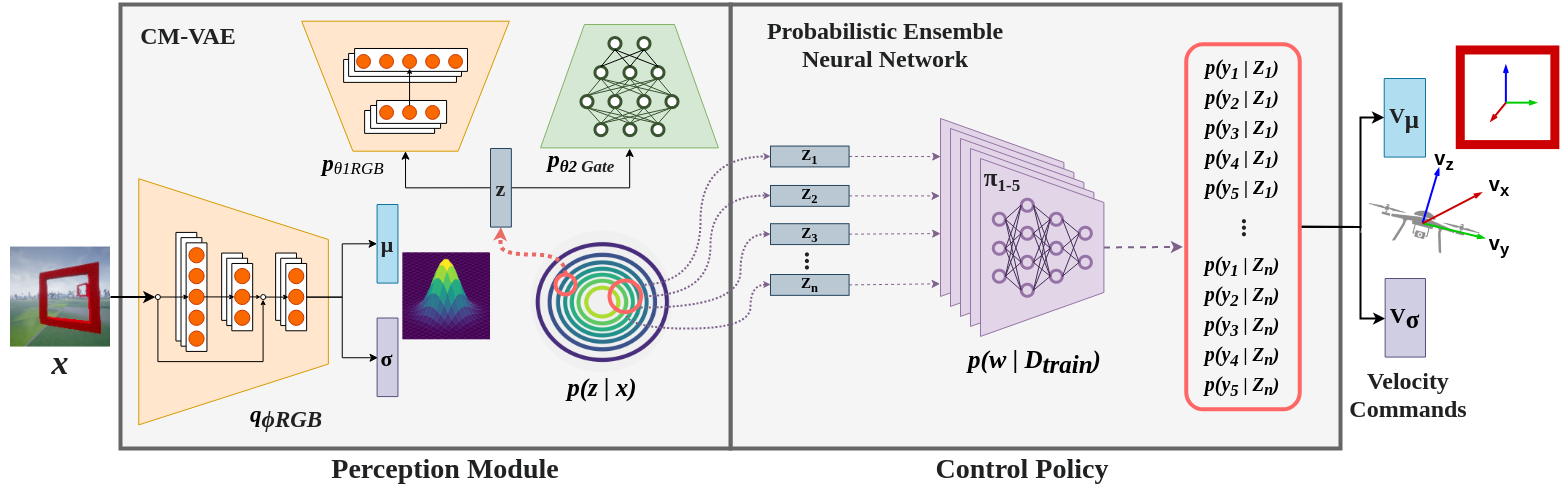}}
		\caption{System Architecture for Autonomous Aerial Navigation.}
		\label{syst-arch}
	\end{center}
	\vskip -0.2in
\end{figure*}

Despite substantial performance improvements introduced by DNNs, they still have significant shortcomings due to their opacity and specially their inability to represent confidence in their predictions. These downsides hinder the deployment of DL methods in safety-critical systems, where uncertainty estimates are highly relevant \cite{arnez2020comparison,loquercio2020general}. To overcome these limitations, the authors in  \cite{mcallister2017concrete} propose the use of Bayesian Deep Learning (BDL) for implementing components used either in modular, E2E, or hybrid fashions. Bayesian methods in DL offer a principled framework to model data (\textit{aleatoric}) and model (\textit{epistemic}) uncertainties to represent the confidence in the outputs.

Nevertheless, system components that use BDL methods should be able to admit uncertainty information as an input, to account for the uncertainty that derives from the outputs of preceding DL-based components. Considering a probability distribution in the input enables the propagation of uncertainty through different components to provide a representative measure of the overall system uncertainty. For example, erroneous or uncertain predictions of an object detector in the perception stage of the pipeline 
can result in an unexpected and unsafe motion plan from the trajectory planner \cite{ivanovic2021heterogeneous}. In a similar fashion, in perception components that learn environmental compact representations (e.g. with Variational Autoencoders \cite{kingma2013auto}), the stochastic nature of the latent space can be use to model complex data-inherent uncertainty like multimodality \cite{depeweg2018decomposition}. In the trajectory planner example, the stochastic latent variables from perception can help to predict multimodal motion plans in ambiguous scenarios like intersections. 

In the literature, only few works such as \cite{henaff2018model,mcallister2019robustness,fan2020learning} consider the use of uncertainty information from other DL components within a system. In this paper, we present a method to account for uncertainty information from stochastic variable inputs that are obtained from other DL components. This method represents a first effort towards handling uncertainty beyond perception in the software pipeline. We evaluate our approach in the context of autonomous aerial navigation, where we aim to learn a robust uncertainty-aware control policy that handles stochastic latent variables from a perception module. To the best of our knowledge, no previous methods handle continuous probability distributions at the inputs of a probabilistic neural network for control in an autonomous navigation task.


\section{Methods}
\subsection{Autonomous Aerial Navigation with DNNs}
In this work, we address the problem of autonomous aerial navigation, in which a UAV aims to traverse a set of gates (with unknown locations) in a drone racing context. To implement our method, we build on the work from \cite{bonatti2019learning}. In this approach, the authors propose a navigation architecture that consists of two components implemented with DNNs: a perception module and a control policy. Figure \ref{syst-arch} depicts the overall system architecture for autonomous navigation.

The objective of the perception module is to extract relevant information from the UAV environment and the current task. To do so, a low dimensional latent representation is learned using a deep generative model. A cross-modal variational autoencoder (CM-VAE) \cite{spurr2018cross} is an effective method to learn a rich, compact, and robust representation. CM-VAEs are variant of traditional VAEs that learn a single latent space with data from multiple sources (modalities). In this case-study, data modalities are represented by RGB images, and the pose of the next gate relative to the UAV body-frame. In the perception module, an RGB image $\mathbf{x}$ at the input is processed by the encoder $q_{\phi RGB}$ into a multivariate normal distribution $\mathcal{N}(\mu, \sigma^2)$ from which latent vectors $z$ are sampled. Later, each data modality can be recovered from the latent space using the decoder $p_{\theta 1 RGB}$ and the estimator $p_{\theta 2 Gate}$. For more details, we refer the reader to \cite{bonatti2019learning,spurr2018cross}.

In the control module, imitation learning is used to train a neural network control policy that maps latent stochastic variables (from CM-VAE) to UAV velocity commands. Different from \cite{bonatti2019learning}, we train an ensemble of probabilistic control policies $\{ \mathbf{\pi}_{i}\}_{i=1}^{M}$ (see Figure \ref{syst-arch}) to account for both, model and data uncertainty. This can be viewed as training a Bayesian Neural Network (BNN) \cite{gustafsson2019evaluating}. In addition, our policy ensemble can handle uncertainty from the latent space (probability distribution) in its input to model complex uncertainty patterns, this is described in detail in the next section. To train the policy ensemble, we employ behavior cloning using raw images with velocity labels. We freeze all the weights from the perception encoder $q_{\phi RGB}$ during the training process.

\subsection{Uncertainty Estimation in DNNs with Stochastic Variable Inputs}
In this work we propose a robust probabilistic control policy model that makes predictions about an output variable $\mathbf{y}$ conditioned on an input $\mathbf{x}$, model parameters $\mathbf{w}$, and a latent variable $\mathbf{z} \sim p(\mathbf{z | x})$. Our control policy model is then represented by the conditional probability distribution $p(\mathbf{y | x, z, w})$. In our case-study, $\mathbf{x}$ represents an RGB image, $\mathbf{y}$ corresponds to UAV velocity commands, and $\mathbf{z}$ represents the low dimensional latent space learned by the encoder model $q_{\phi RGB}(z|x)$.

To capture model uncertainty and handle stochastic inputs (probability distributions), we embrace the Bayesian approach to compute the distribution for a target variable $\mathbf{y^*}$ associated with a new input $\mathbf{x^*}$, using the posterior predictive distribution:

\begin{equation}
	p(\mathbf{y^*|x^*, \mathcal{D}}) = \iint{p(\mathbf{y | z, w}) p(\mathbf{w|\mathcal{D}}) p(\mathbf{z|x^{*}}) \mathbf{dw} \mathbf{dz} }
	\label{eq:postPredDist}
\end{equation}

\begin{algorithm}[tb]
	\caption{Neural Network Predictive Distribution with Stochastic Inputs}
	\label{alg:PredDist-StochasticInputs}
	\begin{algorithmic}
		\STATE {\bfseries Input:} image $\mathbf{x}^*$, policy ensemble $\{ \mathbf{\pi_{w_{i} } \} }_{i=1}^{M} \sim p(\mathbf{w} | \mathcal{D})$
		\STATE Initialize: $\hat{\mu}_{z}$, $\hat{\sigma}_{z}$, $\hat{\mu}_{\pi}$, $\hat{\sigma}_{\pi}$ $\leftarrow$ new empty Tensors
		\STATE Sample: $\{\mathbf{ z}_{n}\}_{n=1}^{N} \sim q_{\phi RGB}(\mathbf{z|x^*})$.
		
		\FOR{$n=1$ {\bfseries to} $N$}

		\FOR{$i=1$ {\bfseries to} $M$}
		\STATE $\hat{\mu}_{\pi}^{\mathbf{w}_i}$, $\hat{\sigma}_{\pi}^{\mathbf{w}_i}$ $\leftarrow$ $\pi_{\mathbf{w}_{i}}(z_{n})$
		\STATE Insert $\hat{\mu}_{\pi}^{\mathbf{w}_i}$ into $\hat{\mu}_{\pi}$; Insert $\hat{\sigma}_{\pi}^{\mathbf{w}_i}$ into $\hat{\sigma}_{\pi}$
		\ENDFOR
		
		\STATE $\hat{\mu}_{z_n}$	$\leftarrow$ $mean(\hat{\mu}_{\pi})$
		\STATE $\hat{\sigma}_{z_n}$ $\leftarrow$ $std(\hat{\sigma}_{\pi}, \hat{\mu}_{z_n}, \hat{\mu}_{\pi})$
		
		\STATE Insert $\hat{\mu}_{z_n}$ into $\hat{\mu}_{z}$; Insert $\hat{\sigma}_{z_n}$ into $\hat{\sigma}_{z}$
		\ENDFOR
		
		\STATE $\hat{\mu}_{\mathbf{y}^*}$	$\leftarrow$ $mean(\hat{\mu}_{z})$
		\STATE $\hat{\sigma}_{\mathbf{y}^*}$ $\leftarrow$ $std(\hat{\sigma}_{z}, \hat{\mu}_{y}, \hat{\mu}_{\pi})$
	\end{algorithmic}
\end{algorithm}

The above integral is intractable, and we rely on approximations to obtain an estimation of the predictive distribution. The distribution $p(\mathbf{w | \mathcal{D}})$ reflects the posterior over model weights given dataset $\mathcal{D}=\{ \mathbf{X, Y, Z} \}$, where $\mathbf{X, Y}$ represent the training inputs and $\mathbf{Z}$ are the latent vectors observed during training. The posterior over the weights is difficult to evaluate, thus we can approximate the inner integral from Equation \ref{eq:postPredDist} using common BNNs methods \cite{arnez2020comparison}. In our case, each member of the ensemble can be viewed as a sample taken from the true posterior distribution over the weights \cite{gustafsson2019evaluating}. Finally, the outer integral is approximated by taking $n$ samples $z_{n}$ from the latent space in our perception model. This last step can be seen as taking a better picture of the perception latent space (with multiple latent samples instead of a single one like in the baseline), leveraging in this way its stochastic nature to take robust control predictions. The overall method is illustrated in Figure \ref{syst-arch}, and summarized in Algorithm \ref{alg:PredDist-StochasticInputs}.

\begin{table}[]
	\caption{Average Number of Gates Traversed by each Aerial Navigation Model (Perception and Control Policy DNNs).}
	\label{table:exp-results}
	\vskip 0.15in
	\begin{center}
		\begin{small}
			\begin{sc}
				\begin{tabular}{@{}lccc@{}}
					\cmidrule(l){2-4}
					\multicolumn{1}{c}{\textbf{}}          & \multicolumn{3}{c}{\textbf{Track Gate Random Noise}} \\ \midrule
					\multicolumn{1}{c}{\textbf{Model}} &
					\begin{tabular}[c]{@{}c@{}}Radius=0\\ Height=0\end{tabular} &
					\begin{tabular}[c]{@{}c@{}}Radius=1 \\ Height=2\end{tabular} &
					\begin{tabular}[c]{@{}c@{}}Radius=1.5\\ Height=2.5\end{tabular} \\ \midrule
					\multicolumn{1}{c}{\textbf{\scriptsize CM-VAE+BC}} & 32   & 12           & 7            \\
					\textbf{\scriptsize CM-VAE+BCE-UI1}                & 32   & 16           & 10           \\
					\textbf{\scriptsize CM-VAE+BCE-UI3}                & 32   & \textbf{20}  & 13           \\
					\textbf{\scriptsize CM-VAE+BCE-UI5}                & 32   & \textbf{20}  & \textbf{17}  \\ \bottomrule
				\end{tabular}
			\end{sc}
		\end{small}
	\end{center}
	\vskip -0.1in
\end{table}

\section{Early Experiments}

For our experiments we used the aerial navigation architecture, the dataset, and the simulation environment from \cite{bonatti2019learning} as baseline. In this work, the architecture is composed by a CM-VAE, and a behavior cloning policy (CM-VAE+BC). To train both components, CM-VAE and BC policy, 300k and 17k RGB images (labeled) are used. In both cases the image size is 64x64 pixels.

We reproduced the baseline (CM-VAE+BC) and implement our method using Pytorch instead of Tensorflow. For the CM-VAE we used the same training settings as in the baseline for the unconstrained version. Different from the baseline, we implemented an ensemble of five neural-networks for our BC control policy (BCE for short). Each member of the ensemble outputs two values, corresponding to the predictive mean and variance, and is trained using the heteroscedastic loss function \cite{kendall2017uncertainties,lakshminarayanan2017simple}. Our BCE policy is able to handle uncertainty at the input according to Algorithm \ref{alg:PredDist-StochasticInputs}, therefore is referred as \textbf{BCE-UI}. In consequence, our aerial navigation architecture for the experiments is referred  as \textbf{CM-VAE+BCE-UIx}, where the last letter indicates the number of samples from the latent space to take into account for the policy predictions.

We evaluate our navigation architecture with the proposed method under controlled simulations that resemble the conditions from the dataset collection. For this purpose, we create a circular track with eight equally spaced gates positioned initially in a radius of 8 m and constant height. To asses the robustness of the navigation policies, we generate new challenging tracks, adding random noise to each gate position (radius) and height. In this way, we force OoD operating conditions.

Table \ref{table:exp-results} shows the average number of gates traversed by each navigation policy under different noise levels for both gate radius (RN) and height (HN). We considered a maximum number of 32 gates which is equivalent to 4 laps in our track. Experimental results show that the navigation policy with our method outperforms the robustness of the other control policies (deterministic and with uncertainty representation only at the output). In fact, the version that includes more input samples is more robust to more drastic OoD conditions. In addition, we find that our navigation policy has a smooth behavior, while the baseline policy and the policy that represents uncertainty only at the output, present a turbulent behavior in noise tracks.

\section{Conclusion}
We presented a method to cope with stochastic variables at the input of BNNs used for control in an aerial navigation task. Based on our experiments, implementing a control policy that accounts for uncertainty at the input improves the robustness of the system in OoD scenarios, besides enabling a smooth behavior. The proposed method represents a step towards uncertainty handling beyond perception in the autonomous system software pipeline. Despite the benefits shown in the experiments, our method relies on sampling, which can be prohibitive in real-time systems. Another direction for future work would be the quality assessment of uncertainty estimates at component and system level, where we should consider the uncertainty propagation capability introduced by our method. This last point represents an important step towards deploying DL-components into safety-critical systems.

%
%
%







\section*{Acknowledgments}
This work has received funding from the COMP4DRONES project, under Joint Undertaking (JU) grant agreement N\degree 826610. The JU receives support from the European Union’s Horizon 2020 research and innovation programme and from Spain, Austria, Belgium, Czech Republic, France, Italy, Latvia, Netherlands.

\bibliography{WAISE_doublecol_paper}
\bibliographystyle{icml2021}

%
%
%

\end{document}